# GCN-TULHOR: Trajectory-User Linking Leveraging GCNs and Higher-Order Spatial Representations

KHOA TRAN*, Lassonde School of Engineering, York University, Canada

PRANAV GUPTA*, Lassonde School of Engineering, York University, Canada

MANOS PAPAGELIS, Lassonde School of Engineering, York University, Canada

Trajectory-user linking (TUL) aims to associate anonymized trajectories with the users who generated them, which is crucial for personalized recommendations, privacy-preserving analytics, and secure location-based services. Existing methods struggle with sparse data, incomplete routes, and inadequate modeling complex spatial dependencies, often relying on low-level check-in data or ignoring intricate spatial patterns. In previous work, we introduced TULHOR [1], a method that transforms raw location data into higher-order mobility flow representations using hexagonal tessellation, effectively reducing data sparsity and capturing richer spatial semantics. In this paper, we propose GCN-TULHOR, an advanced framework that enhances TULHOR by integrating Graph Convolutional Networks (GCNs). Our approach converts both sparse check-in and continuous GPS trajectory data into unified higher-order mobility flow representations, significantly mitigating data sparsity while capturing deeper semantic information. The integrated GCN layer explicitly models complex spatial relationships, capturing non-local dependencies without relying on side information such as timestamps or Points of Interest (POIs). Comprehensive evaluations across six diverse real-world datasets demonstrate consistent improvements over classical baselines, RNN- and Transformer-based models, and the state-of-the-art TULHOR methods in accuracy, precision, recall, and F1-score. Across both sparse and continuous settings, GCN-TULHOR achieves 1–8% relative gains in accuracy and F1-score. Our sensitivity analysis identifies an optimal configuration with a single GCN layer and 512-dimensional embeddings. The integration of GCNs not only enhances spatial learning but also improves the model's generalizability across different types of mobility data. Our work highlights the effectiveness of combining graph-based spatial learning with sequential modeling techniques such as LSTMs, offering a robust and scalable solution for TUL while emphasizing the transferability of the captured spatial knowledge. This advancement has significant implications for personalized recommendations, urban planning, and security, paving the way for more accurate and efficient mobility data analysis. We make our source code publicly available to encourage reproducibility and further research.

---

*Both authors contributed equally to this research.

Authors' Contact Information: Khoa Tran, Lassonde School of Engineering, York University, Toronto, Ontario, Canada, khoasimon99@gmail.com; Pranav Gupta, Lassonde School of Engineering, York University, Toronto, Ontario, Canada, pranavgupta0001@gmail.com; Manos Papagelis, Lassonde School of Engineering, York University, Toronto, Ontario, Canada, papaggel@gmail.com.

---

## 1 INTRODUCTION

**Motivation.** Trajectory-user linking (TUL), the task of attributing anonymous mobility traces to the correct users, is crucial for personalized recommendations, data security, urban planning, epidemiological monitoring, and threat assessments [9, 25]. However, TUL faces significant challenges, as trajectory data is often sparse, noisy, and partially observed, while user mobility patterns are complex and variable [25]. These challenges limit traditional approaches, underscoring the need for methods that effectively capture complex spatial structures and subtle mobility patterns.

**The Significance of Location-Based Services (LBS).** Location-Based Services utilize geographic information from GPS-enabled devices to offer personalized services such as ride-hailing, food delivery, navigation, and targeted advertising [24]. These services depend heavily on analyzing spatiotemporal mobility patterns derived from trajectory data, thus highlighting the critical need for accurate and secure trajectory-user linking methodologies.

**Problem of Interest.** The TUL problem involves associating anonymized trajectory data with corresponding users, framing it as a multi-class classification task. Effective TUL methodologies reveal underlying mobility patterns reflecting individual behaviors, thereby protecting user data, enhancing personalized services, and informing policy-making decisions [9, 14]. Robust solutions must capture the subtleties of daily routines, travel behaviors, and spatial interactions within specific geographic contexts.

**Current Approaches.** Existing TUL methods broadly fall into classical machine learning and deep learning categories. Classical methods leverage trajectory similarity measures such as Dynamic Time Warping (DTW) [3], Longest Common Subsequence (LCSS) [22], and NeuTraj [25]. However, these methods struggle with large, irregular, or noisy datasets. Deep learning approaches—including recurrent neural networks (RNNs), long short-term memory networks (LSTMs) [9], attention mechanisms [14], and variational autoencoders [36]—address these limitations by capturing spatial-temporal dependencies. Nonetheless, these models face issues such as data sparsity, imbalanced user data distributions, and poor generalization across different contexts.

**Limitations of Current Approaches.** Despite advances, current approaches suffer from key limitations:

- **Data Quality:** Trajectory data often exhibits inaccuracies and incompleteness, introducing uncertainty into model predictions.
- **Data Sparsity:** Sparse and irregular trajectory data complicates reliable feature extraction, particularly with raw coordinates or isolated check-ins.
- **Imbalanced Data:** Uneven distribution of trajectories across users complicates fair and accurate classification, potentially biasing outcomes.

Traditional models typically handle location data as raw coordinates or discrete check-ins, failing to capture higher-order spatial dependencies crucial for realistic mobility modeling. Even structured approaches like TULHOR [1], which utilize hexagonal tessellation, treat trajectories as linear sequences, thus overlooking non-local dependencies, indirect routes, and the graph-like structures inherent to urban movement.

**Higher-order Representations (TULHOR).** To address some of these limitations, TULHOR introduced higher-order mobility flow representations, employing geographic tessellation techniques and spatiotemporal embeddings to abstract raw location data [1]. This approach effectively reduced data sparsity and improved trajectory semantics.

The motivation behind this transformation is twofold. First, raw check-in data often captures only isolated events without spatial continuity, while continuous GPS trajectories, although richer in detail, suffer from noise, irregular sampling rates, and redundancy. Representing either of these directly in machine learning models leads to challenges such as high sparsity, poor generalization, and difficulty in capturing latent spatial behaviors. Second, most prior TUL models handle these data types separately, resulting in models that generalize poorly across different urban contexts or data collection modalities.

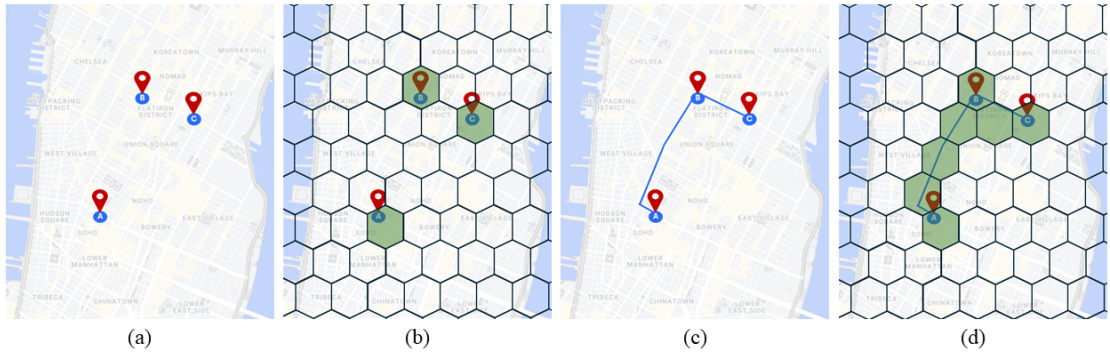

(a)      (b)      (c)      (d)

Fig. 1. An illustrative example showing how sample check-in data from Foursquare-NYC (**a**) can be abstracted to higher-order areas (**b**), and how the sequence of check-ins that infer a trajectory (**c**) can be abstracted to higher-order mobility flow (**d**). We propose GCN-TULHOR, an enhanced model integrating a GCN layer to improve TUL by leveraging higher-order mobility flow representations.

To address these limitations, TULHOR method infers plausible routes between sparse check-ins using routing algorithms (e.g., OSRM), and leverages the full detail of continuous trajectories where available. Both are then discretized into a uniform grid using hexagonal tessellation (following [8]), yielding a graph-structured sequence of grid cells that abstract user movement in a semantically meaningful and topology-aware manner.

This higher-order mobility flow representation offers several key advantages:

- **Unified Representation**: It harmonizes sparse and dense mobility data into a common format, enabling model training and inference on heterogeneous datasets with minimal architectural changes.
- **Reduced Sparsity**: By mapping raw coordinates to hexagonal grid cells, we smooth out gaps in spatial coverage and mitigate the overfitting risk of rare or noisy location points.
- **Enhanced Semantic Structure**: The abstraction reveals latent mobility semantics—such as commonly traversed routes, neighborhood transitions, and regional activity hotspots—that are obscured in raw data.
- **Natural Graph Construction**: The hexagonal grid lends itself to graph-based modeling where each cell becomes a node and adjacency can be defined based on cell connectivity or co-occurrence patterns, which is well-suited for GCNs.
- **Generalizability**: This abstraction layer improves the model's robustness across different cities and data modalities by abstracting over location-specific noise while preserving essential movement dynamics.

This process is illustrated in Figure 1, where check-in data from Foursquare-NYC are transformed into higher-order areas that better reflect underlying mobility semantics. Regardless of whether the original data is sparse or continuous, the resulting representation provides a consistent and spatially informed trajectory view. However, TULHOR still has limited spatial awareness and lacks explicit modeling of spatial relationships between neighboring regions, thus struggling with long-range dependencies and complex movement patterns.

**Distinguishing Visit-Based and Continuous GPS Trajectory Data.** Trajectory classification methodologies differ significantly depending on data collection methods, specifically visit-based versus continuous GPS tracking data. Visit-based data, often from platforms like Foursquare [24], captures sparse, event-triggered check-ins and lacks continuous movement tracking. Consequently, models developed for visit-based data might overfit on sparse, socially motivated location visits and fail to generalize to continuous trajectories. Continuous GPS data, conversely, captures complete movement at regular intervals (e.g., GeoLife [34], T-Drive [29, 30]), allowing models to leverage detailed temporal dynamics and spatial trajectories. This distinction emphasizes the necessity of developing models robust to both sparse and continuous data scenarios.

**Our Approach (GCN-TULHOR).** We propose GCN-TULHOR, a novel framework extending TULHOR by integrating Graph Convolutional Networks (GCNs) to enhance spatial learning. Our method converts both sparse check-ins and continuous GPS trajectories into unified higher-order mobility flow representations via hexagonal tessellation [8], significantly enriching trajectory semantics and reducing data sparsity. By explicitly capturing spatial relationships through graph-based embeddings, GCN-TULHOR addresses TULHOR's limitations by effectively modeling complex spatial dependencies and long-range mobility patterns.

This graph-compatible abstraction enables our model to leverage both temporal patterns and spatial topologies, improving robustness and generalization across heterogeneous datasets. Critically, GCN-TULHOR achieves superior trajectory-user linking without relying on timestamps or Points of Interest (POIs), thus proving highly applicable in privacy-sensitive and data-scarce scenarios.

**Main Contributions.** The key contributions of our work are:

- **Unified Higher-order Mobility Flow Representations:** Enhancing TULHOR by mapping both sparse and continuous trajectory data into hex-based sequences to reduce sparsity and enhance semantics.
- **Spatial Modeling with GCNs:** Introducing a GCN layer to effectively capture spatial embeddings, thus outperforming sequence-based models.
- **Comprehensive Validation:** Demonstrating superior performance of GCN-TULHOR across six diverse datasets in accuracy, precision, recall, and F1-score compared to existing state-of-the-art methods.
- Public availability of our source code to encourage reproducibility and further research.

**Paper Organization.** The rest of the paper is structured as follows: Section 2 details the problem definition, Section 3 describes higher-order mobility flow representations, Section 4 presents the GCN-TULHOR model, Section 5 discusses experimental evaluation, Section 6 explores generalization capabilities, Section 7 reviews related work, and Section 8 concludes the paper.

## 2 PRELIMINARIES AND PROBLEM DEFINITION

This section introduces key definitions and notations, summarized in Table 1, and formally defines the trajectory-user linking problem.

DEFINITION 1 (MAP). *A **map** $\mathcal{M}$ is a finite representation of a geographic area.*

DEFINITION 2 (POINT OF INTEREST (POI)). *A **point of interest** (POI) is a location within the map $\mathcal{M}$ that is of interest to users. The set of all POIs is denoted as $\mathcal{P}$.*

DEFINITION 3 (CHECK-IN RECORD). *A **check-in** (or **visit**) is a record $r = (u, l, t, \langle x, y \rangle)$, where $u$ is a user ID, $l$ is a location ID, $t$ is a timestamp, and $\langle x, y \rangle$ represents the geographic coordinates. The set of all check-in records is denoted by $R$.*

Table 1. Summary of Notations

| Symbol | Description |
|--------|-------------|
| $\mathcal{M}$ | Map of a geographic area |
| $\mathcal{P}$ | Set of points of interest (POIs) in $\mathcal{M}$ |
| $u$ | User ID |
| $l$ | Location ID |
| $t$ | Timestamp |
| $\langle x, y \rangle$ | Geographic coordinates (latitude and longitude) |
| $r$ | Check-in record $(u, l, t, \langle x, y \rangle)$ |
| $Tr$ | Trajectory $Tr = \{r_1, r_2, \ldots, r_m\}$ |
| $\mathcal{G}$ | Set of grid cells from tessellation of $\mathcal{M}$ |
| $g_i$ | The $i$-th grid cell in $\mathcal{G}$ |

DEFINITION 4 (TRAJECTORY). *A **trajectory** $Tr$ is an ordered sequence of a user's check-ins over time: $Tr = \{r_1, r_2, \ldots, r_m\}$, where m is the length of the trajectory.*

**Problem Definition.** Given a set of higher-order mobility flow trajectories $\mathcal{T} = \{Tr_1, Tr_2, \ldots, Tr_n\}$, where each $Tr_i$ is a sequence of grid cells representing a user's mobility flow, and a set of users $\mathcal{U} = \{u_1, u_2, \ldots, u_c\}$, the trajectory-user linking problem is to find a mapping $f : \mathcal{T} \rightarrow \mathcal{U}$ such that $f(Tr_i) = u_i$ for each trajectory.

This is a multi-class classification problem:

$$\min_{f \in \mathcal{F}} \mathbb{E}[\mathcal{L}(f(Tr_i), u_i)], \tag{1}$$

where $\mathcal{F}$ is the space of classifiers, $f$ is a classifier, and $\mathcal{L}(\cdot)$ is a loss function measuring the discrepancy between the predicted user $f(Tr_i)$ and the actual user $u_i$.

By utilizing higher-order mobility flow representations, we enhance the input data for TUL models like GCN-TULHOR, improving its performance.

## 3 HIGHER-ORDER MOBILITY FLOW REPRESENTATIONS

This section introduces higher-order mobility flow representations, explaining their generation from raw trajectory data and their advantages in TUL.

### 3.1 Motivation and Key Idea

Raw check-in data often lacks detailed route information between locations, and GPS data can be noisy and sparse. Directly embedding geographic coordinates in machine learning models is challenging due to their continuous nature.

We transform raw check-in data into **higher-order mobility flow representations** to address these issues. This involves inferring routes between check-ins and abstracting these routes using a grid-based tessellation of the map. Representing trajectories as sequences of grid cells (e.g., hexagons) enriches the data with spatial context and reduces sparsity, enabling models like GCN-TULHOR to better capture user mobility patterns.

### 3.2 Map Tessellation and Grid Representation

Definition 5 (Map Tessellation). *A **map tessellation** divides the map $\mathcal{M}$ into disjoint grid cells $\mathcal{G} = \{g_1, g_2, \ldots, g_n\}$ that cover the area without overlap. Each grid cell $g_i$ is a polygon of uniform size and shape. We use hexagonal cells due to their uniform neighbor relationships and consistent distances.*

Hexagonal tessellations offer benefits over square grids due to the consistent properties of their six neighboring cells, making them better suited for spatial data processing.

### 3.3 Hexagonal Grid Resolutions

In our framework, we follow the Point2Hex method [8] to tessellate the map into a grid of regular hexagonal cells, forming the basis for spatial abstraction in higher-order trajectory representations. The tessellation is performed at multiple levels of spatial granularity, each defined by a resolution level (e.g., Hex6, Hex7, ..., Hex10). These resolutions control the size and shape of each hexagonal cell, thereby impacting both the expressiveness and computational complexity of the model.

Each increase in resolution leads to finer-grained tessellation, allowing for more precise modeling of spatial dynamics, but also introduces additional sparsity. Table 2 provides the characteristics of hexagons at each resolution, as presented in the Point2Hex study.

Table 2. Hexagon size characteristics at different resolutions [8].

| Resolution | Edge Length (km) | Area (km$^2$) |
|---|---|---|
| Hex@6 | 3.725 | 36.129 |
| Hex@7 | 1.406 | 5.161 |
| Hex@8 | 0.531 | 0.737 |
| Hex@9 | 0.201 | 0.105 |
| Hex@10 | 0.076 | 0.015 |

Higher resolutions such as Hex@9 and Hex@10 are especially suitable for urban mobility studies where fine spatial variations are essential, while lower resolutions (e.g., Hex@6) are beneficial for macroscopic patterns over larger geographies. In GCN-TULHOR, we empirically evaluate and optimize these resolution levels to balance trajectory coverage, computational tractability, and sparsity reduction.

Furthermore, using hexagons over squares or triangles mitigates the directional bias introduced by grid shapes and ensures uniform neighborhood relationships, which improves learning performance in spatial GCN models.

### 3.4 Generating Higher-order Mobility Flow Data

The process of generating higher-order mobility flow data includes the following steps:

(1) **Route Estimation and Map-Matching**:
  - Routes between consecutive check-in locations are estimated using routing algorithms (e.g., OSRM [13]).
  - Map-matching techniques align routes with the road network to correct GPS inaccuracies [16].

(2) **Higher-order Check-ins and Trajectories**:

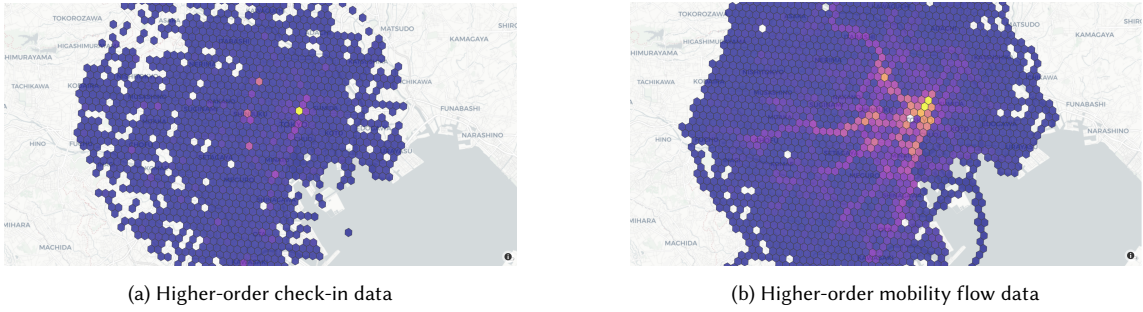

(a) Higher-order check-in data                    (b) Higher-order mobility flow data

Fig. 3. An example (FOURSQUARE-NYC) showing the transformation from higher-order check-in data to higher-order mobility flow data, enriching trajectory semantics and helping address the TUL problem.

DEFINITION 6 (HIGHER-ORDER CHECK-IN). *Each check-in location $p \in \mathcal{P}$ is mapped to the grid cell $g \in \mathcal{G}$ that contains it.*

DEFINITION 7 (HIGHER-ORDER TRAJECTORY). *A trajectory $Tr = \{p_1, p_2, \ldots, p_m\}$ is transformed into a sequence of grid cells $Tr = \{g_1, g_2, \ldots, g_m\}$, where each $g_i$ is the grid cell containing $p_i$.*

(3) **Higher-order Mobility Flow**:

DEFINITION 8 (HIGHER-ORDER MOBILITY FLOW). *For each route between $p_i$ and $p_{i+1}$, we determine the sequence of traversed grid cells. The higher-order mobility flow is the concatenation of these sequences, resulting in $Tr = \{g_1, \langle g_{1,2} \rangle, g_2, \langle g_{2,3} \rangle, \ldots, \langle g_{m-1,m} \rangle, g_m\}$, where $\langle g_{i,i+1} \rangle$ represents the sequence of grid cells between $g_i$ and $g_{i+1}$.*

Figure 3 shows how higher-order mobility flow data enriches trajectory semantics, revealing road networks and frequently traveled paths not apparent in raw check-in data.

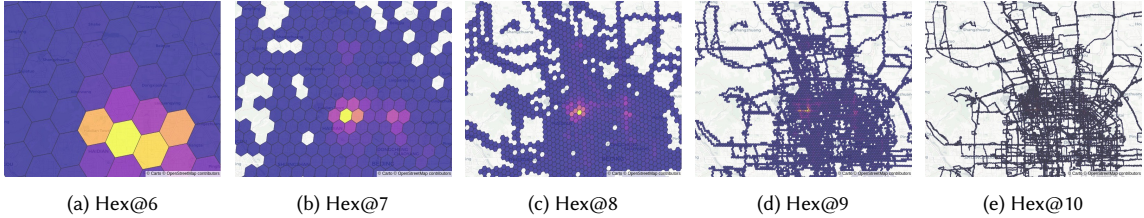

(a) Hex@6          (b) Hex@7          (c) Hex@8          (d) Hex@9          (e) Hex@10

Fig. 2. Visualization of HO-Geolife trajectory data at different hexagonal resolutions. As resolution increases (from Hex@6 to Hex@10), the tessellation becomes finer, capturing more detailed movement patterns while also increasing spatial granularity.

## 3.5 Advantages in Addressing the TUL Problem

Higher-order mobility flow representations offer several benefits for TUL:

- **Reduced Sparsity**: Aggregating check-ins and routes into grid cells reduces data sparsity, as multiple POIs and routes contribute to the same grid cell. Figure 4 shows this impact for benchmark datasets.
- **Enhanced Mobility Patterns**: This representation captures detailed mobility flows, providing richer information about user movement patterns.

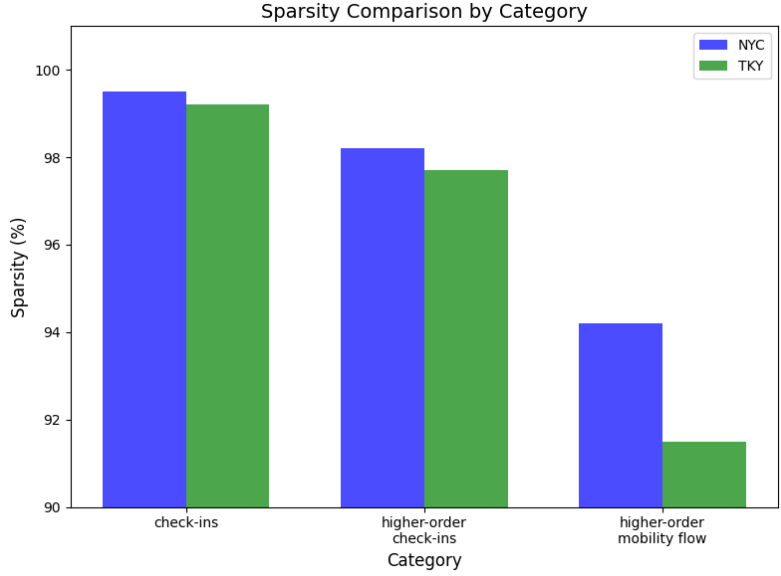

Fig. 4. Impact of higher-order abstraction on data sparsity for FOURSQUARE-NYC and FOURSQUARE-TKY datasets.

- **Improved Model Performance**: Models like GCN-TULHOR can better distinguish between users based on their mobility behaviors using this structured data.
- **Scalability and Flexibility**: The grid-based representation allows for varying levels of resolution, offering a trade-off between granularity and computational efficiency.

### 3.6 Addressing the Sparsity Problem

Check-in and mobility trajectory data are inherently *sparse* due to the irregular and infrequent nature of user activity. This sparsity is typically reflected in the *user-location interaction matrix*, where an entry $(i, j)$ is marked as 1 if user $i$ has visited location $j$, and 0 otherwise. In real-world datasets, this matrix contains a large proportion of zeros, indicating that users only visit a small subset of all possible locations. Furthermore, the distribution of visits is often highly *skewed* — a few locations receive the majority of check-ins, while the vast majority remain underrepresented.

This high level of sparsity poses a significant challenge for trajectory-user linking and other downstream modeling tasks. Sparse interaction patterns limit the learning capacity of models and may lead to biased or incomplete representations of user mobility. Additionally, directly embedding latitude and longitude coordinates into machine learning models is problematic due to their continuous nature and fine-grained resolution, which often fails to generalize across users.

To mitigate these issues, we adopt a strategy inspired by *higher-order spatial representations*. Instead of modeling interactions at the raw location level, we aggregate them into coarser spatial units, such as *hexagonal grid cells*. This transforms the sparse user-location matrix into a *denser user-grid cell matrix*, where each entry indicates whether a user has visited any location within a given spatial cell. This aggregation offers two major benefits:

(1) **Reduces sparsity** by broadening the scope of each interaction, allowing multiple locations to be merged into a single spatial unit.
(2) **Enhances generalization** by capturing higher-level mobility patterns that are more robust across users and regions.

For example, consider a scenario in which three users each visit a single POI. In a $3 \times 3$ user-POI matrix, the sparsity would be 66.7%. If two of those POIs fall into the same grid cell, the equivalent user-grid cell matrix would be $3 \times 2$ with the same number of positive entries, reducing sparsity to 50%. Empirically, we observe that applying higher-order representations consistently decreases sparsity across datasets, facilitating improved learning and model performance.

## 3.7 Technical Considerations

*Map-Matching.* Map-matching is needed to align routes with the road network, correct GPS errors, and ensure realistic trajectories. Hidden Markov Model-based algorithms [16] or routing services like OSRM [13] can be used.

*Computational Geometry.* Transforming routes into grid cell sequences involves intersecting linestrings (routes) with polygons (grid cells), which are facilitated by efficient algorithms and libraries (e.g., `shapely`, `geopandas`).

## 4 METHODOLOGY

In this section, we are expanding the capabilities of TULHOR, a spatiotemporal learning framework inspired by BERT, which leverages higher-order mobility flow representations for trajectory-user linking. However, TULHOR has shown some limitations mentioned in (add ref to related)

To overcome these limitations, we propose GCN-TULHOR, an enhanced framework that integrates Graph Convolutional Networks (GCNs) into the TULHOR pipeline. By leveraging graph-based spatial learning, GCN-TULHOR effectively models both local and global movement dependencies, allowing for a more robust, scalable, and context-aware trajectory-user linking approach. This integration addresses TULHOR's constraints by introducing spatially-aware embeddings, enabling the model to capture complex mobility patterns beyond what is possible with sequence-based learning alone.

One of their key benefits is that edges can be defined dynamically based on the dataset, rather than being restricted to predefined topologies such as road networks or fixed spatial grids. This data-driven edge formulation allows GCN-TULHOR to model spatial relationships adaptively, capturing movement patterns, trajectory overlaps, and mobility behaviors more effectively. Additionally, GCNs aggregate information from neighboring nodes, ensuring that both local and non-local dependencies are incorporated into the learned embeddings. This makes GCN-TULHOR highly generalizable across diverse datasets, whether dealing with sparse check-ins or continuous GPS trajectories.

By integrating GCN-based spatial learning with higher-order mobility flow representations, GCN-TULHOR achieves a more robust, scalable, and accurate trajectory-user linking model compared to traditional graph methods, making it well-suited for real-world mobility analytics applications.

## 4.1 TULHOR's Spatiotemporal Embedding Layer and GCN cooperation

The spatial-temporal embedding layer converts sparse one-hot encodings of check-in components (grid cells, POIs, and timestamps) into dense vector representations. POI information is included to differentiate mobility patterns that traverse the same grid cell sequence. For example, Alice and Bob might have identical trajectories across campus grid cells, but if they frequent different POIs (e.g., lecture halls vs. labs), their semantic patterns differ. The embedding process is formalized as:

$$z_i^g = \phi_g(g_i, W_g) \tag{2}$$

$$z_i^p = \phi_p(p_i, W_p) \tag{3}$$

$$z_i^s = \text{GCN}(g_i, \mathcal{G}) \tag{4}$$

Here, $g_i$ is the grid cell ID, $p_i$ is the POI visited within $g_i$, and $z_i^g$, $z_i^p$, and $z_i^s$ represent the embeddings for grid cell semantics, POI identity, and spatial context, respectively. The first two embeddings, $z_i^g$ and $z_i^p$, are produced by standard embedding layers $\phi_g(.)$ and $\phi_p(.)$ using learnable parameters $W_g$ and $W_p$, which are randomly initialized and trained jointly. In contrast, the spatial embedding $z_i^s$ is learned via a Graph Convolutional Network, which operates on the trajectory-induced graph $\mathcal{G}$. Each node in $\mathcal{G}$ represents a hexagonal grid cell, and edges capture adjacency or mobility flow patterns between cells.

This GCN-based spatial embedding replaces the static spatial layer in the original TULHOR framework. Rather than initializing $W_s$ and freezing it, we dynamically learn spatial features based on neighborhood relationships, enabling the model to capture richer, non-local spatial dependencies. All embeddings have a consistent dimensionality $d_L$. Specifically, $W_g \in \mathbb{R}^{n \times d_L}$, $W_p \in \mathbb{R}^{n_p \times d_L}$, where $n$ is the number of grid cells and $n_p$ the number of POIs.

The timestamp $t_i$ is a continuous feature, and therefore, regarding it directly as an input feature will lead to a loss of information since the embeddings will not scale linearly in the feature space. The aim is to learn timestamp embeddings that preserve the properties of time, such as periodicity. Furthermore, the distance in the embedding space between two timestamps needs to be proportional to the difference between the timestamps, i.e., the relative information between the timestamps must be preserved. Inspired by existing work [11], we design a temporal-aware positional encoding to replace the positional encoding used in the original BERT model with:

$$[z_i^t]_j = \begin{cases} \sin(w_j t_i), & \text{if } j \text{ is odd} \\ \cos(w_j t_i), & \text{if } j \text{ is even} \end{cases} \tag{5}$$

where $j$ is the order of the dimension, $w_j$ is a learnable parameter, and $t_i$ is the timestamp of the $i$th check-in in the trajectory. To see why this temporal encoding preserves the relative information between the timestamps, we can calculate the distance between two consecutive timestamps as:

$$(z_i^t)(z_{i+1}^t)^\top = \sum_{i=1}^{d} \cos(w_i(t_{i+1} - t_i)) \tag{6}$$

where the distance between $t_i$ and $t_{i+1}$ timestamps is the dot product of their respective temporal encoding $(z_i^t)$ and $(z_{i+1}^t)$. We can observe that the distance between the vectors is dependent on the difference between the timestamps $t_{i+1} - t_i$ and on $w_i$ (parameters which the model learns during the training). Thus, the relative and periodic information of time is preserved and learned in this encoding function.

We adopt a non-invasive self-attention mechanism [12] where the side information, like spatial and temporal properties, is passed to the self-attention module directly instead of adding it to the grid cell embeddings. Therefore, the spatial-temporal embedding layer produces two outputs:

$$R^{(id)} = z_1^g, z_2^g, ..., z_m^g \tag{7}$$

$$R = (\{z_1^s, z_2^s, ..., z_m^s\}, \{z_1^p, z_2^p, ..., z_m^p\}, \{z_1^t, z_2^t, ..., z_m^t\}) \tag{8}$$

where $R^{(id)}$ embeddings are passed forward to the encoder, while the $R$ embeddings are passed directly to the self-attention component, as shown in the Figure 6. The $R^{(id)}$ contains the embeddings of the grid cells, while $R$ contains three sets, each one having the embedding of different side information like spatial, temporal, and POIs.

## 4.2 GCN-TULHOR's Encoder

The encoder block consists of *a multi-head spatial-temporal non-invasive self-attention* mechanism followed by *a position-wise feed-forward layer*. The self-attention enriches each token with *spatial*, *temporal*, and *contextual* information from other tokens in the sequence. For the model to attend to all these different dependencies, the self-attention mechanism uses multiple heads, allowing the model to capture various dependencies in parallel. Following the multi-head self-attention component, there is a position-wise feed-forward network with ReLu activation function to introduce non-linearity. Next, there is a residual connection, which allows the gradient to flow through the model without exploding or vanishing, making the training stable. The training is further stabilized using layer normalization.

Note that the multi-head spatial-temporal non-invasive self-attention (ST-NOVA) in GCN-TULHOR differs from the standard self-attention (SA) found in Transformer models. The standard self-attention is represented as:

$$SA(Q, K, V) = \sigma(\frac{QK^T}{\sqrt{d_n}})V \tag{9}$$

where $Q, K, V \in \mathbb{R}^{m \times d_n}$, $d_n$ is the hidden state embedding dimensions, and $\sigma$ is the softmax operation. *SA* calculates the weighted average for each token in the sequence based on its corresponding similarity with the other tokens in the same sequence. *SA* uses an invasive-attention, implying that any additional features, such as positional information, must be infused into the input sequence representation. This has a major drawback because the output of the self-attention layer is fed to the predicting layer, which tries to search the token ID space; if we were to add additional features to the sequence representation, then we would end up creating a compounded embedding space, which makes the searching task harder. To address these problems, we use a non-invasive attention instead, which is represented as:

$$\text{ST-NOVA}(R^{(id)}, R) = \sigma\left(\frac{QK^\top}{\sqrt{d_L}}\right)V \tag{10}$$

$$V = R^{(id)} \times W_V, \quad K = F \times W_K, \quad Q = F \times W_Q \tag{11}$$

$$F = \text{MLP}(R^{(id)} \parallel R) \tag{12}$$

where $W_V, W_k, W_Q \in \mathbb{R}^{d_L \times d_n}$, $F \in \mathbb{R}^{m \times d_L}$ and $\parallel$ is the concatenation operation. The ST-NOVA takes two inputs, the input sequence id $R^{(id)}$ and the other side information in $R$. Then ST-NOVA uses the input sequence id $R^{(id)}$ to calculate the Values matrix. Regarding the Keys and Query matrices, the component concatenates the input sequence id with the additional features and uses a multilayer perceptron (MLP) to unify the dimension; the output of the MLP is used to calculate the Keys and Query matrices. ST-NOVA uses the additional features to calculate how tokens are similar. Unlike *SA*, which infuses the additional features directly into the input sequence, we use the additional features to understand how two tokens are similar.

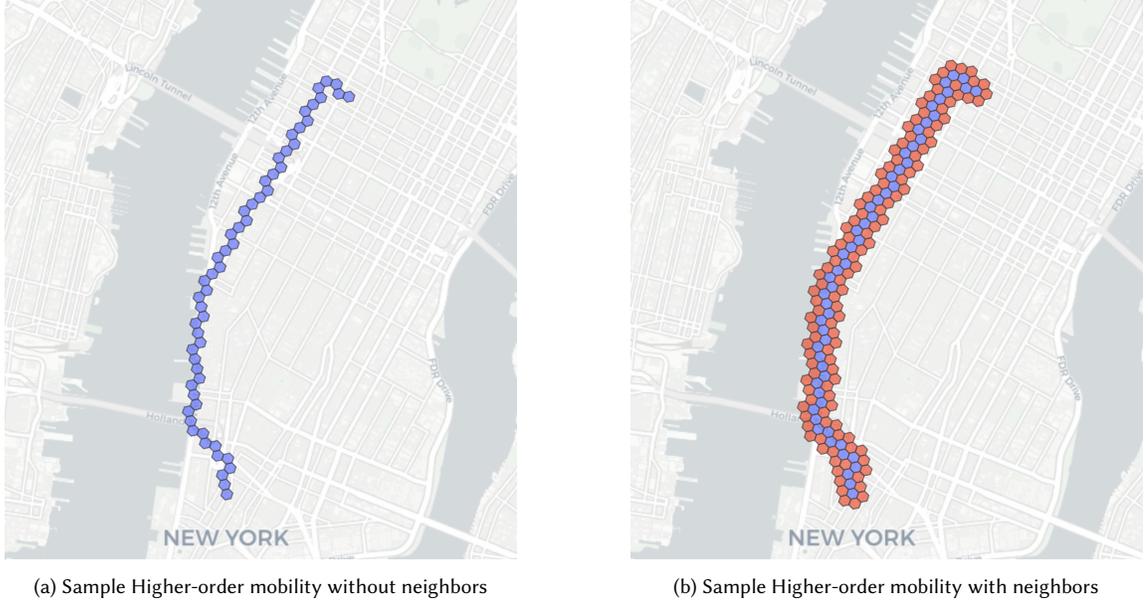

(a) Sample Higher-order mobility without neighbors        (b) Sample Higher-order mobility with neighbors

Fig. 5. An example (Foursquare-NYC-Continuous) showing how trajectory neighbors are leveraged in GCN-TULHOR

## 4.3 Pre-training and Fine-tuning GCN-TULHOR

The model is pre-trained using a masked language modeling (MLM) objective:

$$\mathcal{L}_{MLM} = \frac{1}{|Tr^m|} \sum_{g_m \in Tr^m} -\log P(g^m = g^{m^*} | Tr^{m'}), \tag{13}$$

with 40% of tokens masked to accelerate convergence. The model is fine-tuned for trajectory-user linking by adding a classification layer using a balanced cross-entropy loss:

$$\mathcal{L}(Tr, u_i) = \frac{1 - \beta}{1 - \beta^{n_{u_i}}} \log(\sigma(y')), \tag{14}$$

where $\beta$ controls the balancing factor, and $n_{u_i}$ is the count of trajectories for user $u_i$.

This methodology captures realistic trajectory-user linking challenges, reflecting continuous user movement while addressing data sparsity and path overlap issues.

## 4.4 GCN for Spatial Embedding Layer

This section details our extension of the TULHOR model with a GCN layer, focusing on the architectural enhancements, design rationale, and expected improvements. The high-level architecture of our proposed framework is illustrated in Figure 6.

*4.4.1 Integration of GCN Layer.* To enhance the spatial learning capabilities of TULHOR, we integrated a GCN layer, which excels in learning from graph-structured data. Figure 5 demonstrates how trajectory neighbors in higher-order flows are utilized by GCN-TULHOR to capture complex spatial relationships. This layer captures spatial relationships by operating on graphs where nodes represent geographic locations (e.g., hexagonal grid cells or POIs), and edges represent the connections between these locations. Each node is characterized by a feature vector derived from trajectory data, and the GCN learns node embeddings that encapsulate spatial dependencies.The GCN layer's output embeddings enhance the model's spatial context and are combined with temporal and contextual embeddings from the original TULHOR framework.

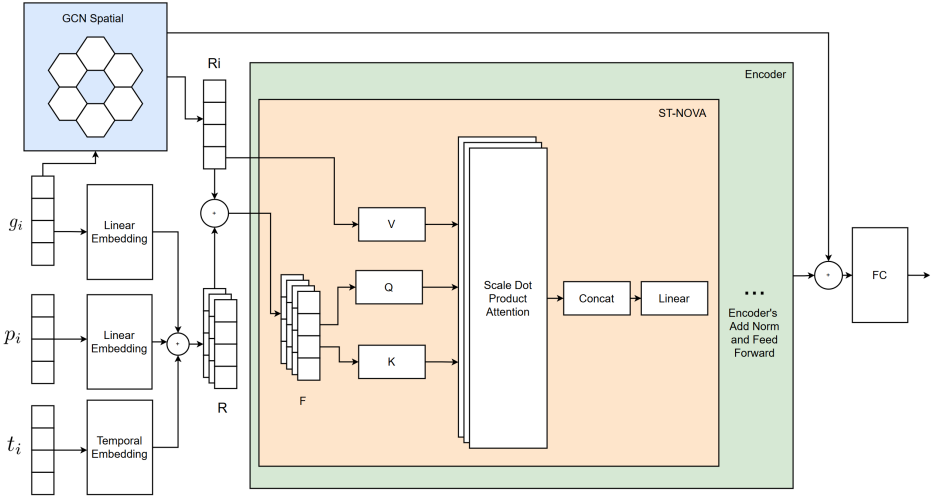

Fig. 6. High-level architecture of GCN-TULHOR.

*4.4.2 Graph Construction.* The spatial structure is represented by a graph $G = (V, E)$, where each node $v_i \in V$ corresponds to a hexagonal grid cell, and each edge $e_{ij} \in E$ represents a spatial relationship between adjacent grid cells. Nodes include all observed hexagonal cells, and the embedding layer learns representations for each cell.

*4.4.3 Node Embeddings.* Each node $v_i$ in $G$ starts with an initial embedding $\mathbf{h}_i^{(0)} \in \mathbb{R}^d$, where $d$ is the embedding dimension. The embedding layer maps each grid cell to a $d$-dimensional space, enabling the learning of spatial representations. These embeddings are updated as the GCN aggregates information from neighboring nodes.

*4.4.4 Neighborhood Aggregation.* The GCN aggregates spatial information by updating node embeddings based on their neighbors. Instead of simple averaging, we introduce the adjacency matrix $A$ to weigh the contributions from neighboring nodes dynamically. The aggregation process for the next layer $l + 1$ is now defined as:

$$\mathbf{h}_i^{(l+1)} = \sigma \left( \sum_{j \in \mathcal{N}(i)} A_{ij} \mathbf{W}^{(l)} \mathbf{h}_j^{(l)} \right),$$ (15)

where: - $\mathbf{h}_i^{(l)}$ is the node embedding at layer $l$, - $\mathbf{W}^{(l)}$ is a learnable weight matrix, - $A_{ij}$ is the normalized adjacency matrix, determining the influence of node $v_j$ on node $v_i$, - $\mathcal{N}(i)$ represents the neighbors of $v_i$, - $\sigma$ is an activation function, such as ReLU.

*4.4.5 Adjacency Matrix Definition.* Let $V$ be the set of all hexagonal cells in the spatial grid and $\mathcal{T}$ the set of all observed trajectories, each represented as an ordered list of visited hexagon indices. For each node $v_i \in V$, let $\mathcal{N}_{obs}(i) \subset V$ be the set of neighboring hexagons that directly follow $v_i$ in any trajectory, and $\mathcal{N}_{geo}(i) \subset V$ be the set of geometrically adjacent hexagons (up to 6 due to hexagonal tessellation).

We define the total set of neighbors for node $v_i$ as:

$$\mathcal{N}(i) = \mathcal{N}_{obs}(i) \cup \left( \mathcal{N}_{geo}(i) \setminus \mathcal{N}_{obs}(i) \right), \tag{16}$$

where: - $\mathcal{N}_{obs}(i)$ are data-driven (observed in trajectories), - $\mathcal{N}_{geo}(i) \setminus \mathcal{N}_{obs}(i)$ are structural (adjacent due to hex shape but not seen in data).

The raw weight between any pair $(v_i, v_j)$ is then defined as:

$$\text{Raw}_{ij} = \begin{cases} \text{Count}(v_i, v_j) & \text{if } v_j \in \mathcal{N}_{obs}(i), \\ 1 & \text{if } v_j \in \mathcal{N}_{geo}(i) \setminus \mathcal{N}_{obs}(i), \\ 0 & \text{otherwise.} \end{cases} \tag{17}$$

We then perform row-normalization over all neighbors:

$$A_{ij} = \frac{\text{Raw}_{ij}}{\sum_{k \in \mathcal{N}(i)} \text{Raw}_{ik}}. \tag{18}$$

This ensures that all immediate neighbors (observed or geometric) of a node contribute to its representation. The observed transitions dominate when supported by data, while geometric adjacency provides a smoothing prior to account for unobserved but spatially close regions.

For Graph Convolutional Network (GCN) processing, we further apply symmetric normalization and introduce self-loops:

$$\tilde{A} = D^{-\frac{1}{2}} (A + I) D^{-\frac{1}{2}}, \tag{19}$$

Unlike traditional adjacency matrices that rely on predefined topologies and uniform grid structures, the adjacency matrix in GCN-TULHOR is computed dynamically from the trajectory dataset itself. This allows the graph structure to evolve based on real-world movement patterns rather than being constrained by a fixed spatial representation.

The ability to define edges in a data-driven manner makes the model highly adaptable. Instead of relying on predefined networks, the adjacency matrix adjusts dynamically based on the frequency of user transitions between different locations. This ensures that the connectivity structure reflects actual user mobility behaviors, allowing GCN-TULHOR to capture both common and less frequent movement patterns.

Additionally, this formulation enhances spatial representation by assigning greater importance to high-traffic regions. Areas with frequent transitions between locations receive stronger connections, reinforcing movement pathways that play a crucial role in user trajectory prediction. By doing so, the model effectively distinguishes between primary transit routes and less significant movement patterns, improving classification accuracy.

Another advantage of this approach is its scalability. Since the adjacency matrix is inferred from observed data, it can flexibly scale across different geographic regions and datasets without requiring manual adjustments. This makes GCN-TULHOR applicable to a wide range of urban and mobility datasets, ensuring that the model remains effective across diverse environments.

By leveraging a data-driven adjacency matrix, the GCN framework efficiently captures complex mobility relationships, providing a more accurate and context-aware trajectory-user linking approach.

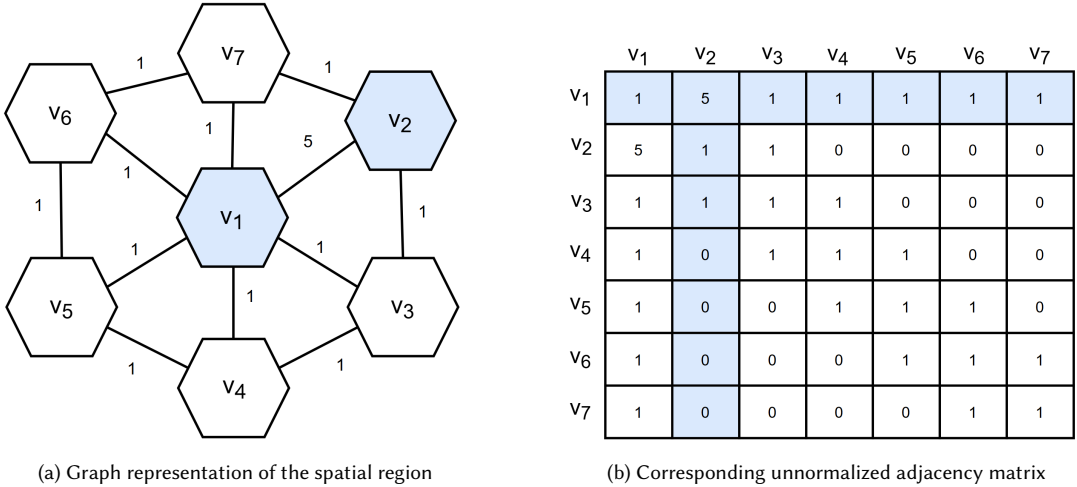

(a) Graph representation of the spatial region    (b) Corresponding unnormalized adjacency matrix

Fig. 7. An illustrative example of how the adjacency matrix $A$ is constructed. The left diagram shows a portion of a hexagonally tessellated space, where nodes $v_1$ and $v_2$ (in blue) are hexagonal regions observed in the trajectory dataset. All other hexagons are geometrically adjacent but do not appear in the data. For each node, a self-loop is included (i.e., $A_{ii} = 1$), and a weight of 1 is assigned to each geometric neighbor that was not observed in a trajectory. Data-driven transitions (e.g., $v_1 \rightarrow v_2$) are weighted by their actual frequency (in this case, 5).

*4.4.6 GCN Layer and Skip Connection.* The GCN layer processes the graph and outputs updated node embeddings. A skip connection is introduced from the GCN output directly to the prediction layer for efficient training and gradient flow. This allows the model to leverage both GCN-based spatial features and high-level features from the Transformer encoder. The final prediction layer combines outputs from both the GCN and Transformer components for trajectory-user linking.

## 5 EVALUATION

In this Evaluation section, we assess GCN-TULHOR by benchmarking it against state-of-the-art models on diverse real-world datasets. These include both sparse check-in datasets (e.g., Foursquare NYC and Tokyo) and continuous trajectory datasets (e.g., T-Drive, Porto, Rome, and Geolife), offering a wide spectrum of mobility behaviors.

To ensure a consistent and expressive input representation across data types, all datasets are transformed into higher-order mobility flows (HO) using hexagonal tessellation. For sparse check-ins, plausible routes are inferred between POIs, while continuous GPS traces are discretized into hex-cell sequences to preserve trajectory semantics. This unified HO representation allows GCN-TULHOR to effectively model both types of mobility data within a common spatial framework.

We then compare GCN-TULHOR with both traditional machine learning models (e.g., Decision Trees, SVMs) and deep learning approaches (e.g., RNNs, LSTMs, TULHOR). Through metrics like accuracy, precision, recall, and F1-score, we demonstrate its superior trajectory-user linking performance across both sparse and continuous datasets.

Next, we analyze how GCN integration improves spatial learning and generalization, followed by a hyperparameter study on GCN layers, embedding size, and grid resolution (HEX levels) to optimize performance. This evaluation confirms GCN-TULHOR as a scalable, robust, and spatially-aware solution for trajectory-user linking.

## 5.1 Evaluation Data

We use two categories of data to evaluate the TUL performance of GCN-TULHOR:

*5.1.1 Check-in Datasets.* We use two real-world datasets from Foursquare[1] social network: New York (NYC) and Tokyo (TKY) from 2012–2013 [24]. We filter out trajectories with fewer than three check-ins and users with fewer than five trajectories. The data is split into 80% training and 20% test sets. We applied the tessellation process describe in Point2Hex, forming **HO-Foursquare-NYC-Check-ins** (abbreviated as HO-NYC-CI) and **HO-Foursquare-TKY-Check-ins** (abbreviated as HO-TKY-CI) datasets.

These datasets evaluate model performance under sparse conditions, where continuous movement information is limited. Contrasting performance with continuous datasets reveals GCN-TULHOR's robustness and adaptability.

*5.1.2 Continuous Trajectory Datasets.* To assess generalization, we use six datasets with diverse geographic regions, sampling frequencies, and mobility characteristic. Each dataset is transformed into higher-order mobility flows, converting raw location data into hexagon-based trajectories. We remove auxiliary information like timestamps and POI categories to focus on spatial structure. These datasets test GCN-TULHOR under conditions like overlapping routes, variable sampling densities, and complex mobility:

- **HO-T-Drive** [29, 30]: Taxi trajectories from Beijing, with about 9 million km of driving data. This challenges models to maintain accuracy over large areas and long distances.
- **HO-Porto** [15]: Trajectories of Portuguese taxis with a dispatch system, offering insight into structured taxi routing and service-oriented mobility.
- **HO-Rome** [4]: Taxi cab traces from Rome, Italy, providing a European urban context.
- **HO-Geolife** [33–35]: A GPS-based dataset containing trajectories of various outdoor activities over about 1.2 million km, testing the model's ability to handle diverse behaviors.
- **HO-Foursquare** (NYC and Tokyo): POI check-ins transformed into continuous trajectories via higher-order mobility flows and tessellation, referred to as **HO-Foursquare-NYC-Continuous** (abbreviated as NYC-CON) and **HO-Foursquare-TKY-Continuous** (abbreviated as TKY-CON).

**Dataset Statistics** Table 3 summarizes key statistics for check-in and continuous datasets at a chosen hexagonal resolution (Hex8), reporting:

- $|\mathcal{T}|$: Total number of trajectories.
- $|\mathcal{U}|$: Total number of unique users.
- $|\mathcal{T}_{\text{uniq}}|$: Number of unique trajectories.
- $|\mathcal{H}_{\text{uniq}}|$: Number of unique hexagonal cells.

## 5.2 Baselines and Implementation

*5.2.1 Baselines for Check-in Data.* We compare GCN-TULHOR with:

- **TULHOR**: A BERT-inspired spatiotemporal learning framework that uses higher-order mobility flow representations for trajectory-user linking.
- **Classical ML methods**: Decision Tree (DT), Linear Discriminant Analysis (LDA), and Linear Support Vector Machine (SVM). We used Bag-of-Words (BOW) with Singular Value Decomposition (SVD).

---

Table 3. Dataset Statistics for HEX8.

| Dataset | $|\mathcal{T}|$ | $|\mathcal{U}|$ | $|\mathcal{T}_{\mathbf{UNIQ}}|$ | $|\mathcal{H}_{\mathbf{UNIQ}}|$ |
|---|---|---|---|---|
| **HO-NYC-CI** $|\mathcal{U}|$=**108** | 6,489 | 108 | 6,489 | 3,197 |
| **HO-NYC-CI** $|\mathcal{U}|$=**209** | 9,200 | 209 | 9,200 | 4,098 |
| **HO-NYC-CI** $|\mathcal{U}|$=**234** | 9,671 | 234 | 9,671 | 4,287 |
| **HO-TKY-CI** $|\mathcal{U}|$=**108** | 8,710 | 108 | 8,710 | 980 |
| **HO-TKY-CI** $|\mathcal{U}|$=**209** | 13,292 | 209 | 13,292 | 1,232 |
| **HO-TKY-CI** $|\mathcal{U}|$=**451** | 19,549 | 443 | 19,549 | 1,374 |
| **HO-NYC-CON** | 49,983 | 1,083 | 34,903 | 2,318 |
| **HO-TKY-CON** | 117,593 | 2,293 | 73,266 | 1,933 |
| **HO-Porto** | 1,668,859 | 442 | 552,268 | 12,998 |
| **HO-Rome** | 5,873 | 315 | 5,871 | 875 |
| **HO-Geolife** | 2,100 | 57 | 2,006 | 6,360 |
| **HO-T-Drive** | 65,117 | 9,987 | 64,857 | 25,745 |

- **TULER** [9]: An RNN-based TUL model with RNN, LSTM, and GRU variations.
- **DeepTUL** [14]: An RNN with attention, evaluated with RNN, LSTM, and GRU variations.

*5.2.2 Baselines for Continuous Data.* We compare GCN-TULHOR with:

- **TULHOR**: A sequence-based TUL method leveraging geographic tessellation and spatiotemporal embeddings to improve trajectory representation learning.
- **Classical ML methods**: Decision Tree (DT), Linear Discriminant Analysis (LDA), and Linear Support Vector Machine (SVM) with Bag-of-Words (BOW) and SVD.
- **Recurrent Neural Networks (RNNs)**: Traditional RNN, LSTM, and GRU models.

**Reasons for removing TULER and DeepTUL from Continuous Dataset:** TULER[9] and DeepTUL[14] are primarily designed for check-in data and rely heavily on temporal features like sequential time series. Their architectures are optimized for discrete check-in events, making them unsuitable for modeling continuous trajectories, which require a different approach to handle fine-grained user movement over time.

**Implementation**: GCN-TULHOR was implemented in PyTorch, using a single encoder layer with 12 attention heads. For GCN-TULHOR, the GCN layer was set to 1, and the GCN embedding size was 256. Both models used a batch size of 24, a learning rate of 0.005, a learning rate decay of 0.5, and an embedding size of 512. The momentum parameter $\beta$ was set to 0.99. Training was performed for 10 epochs. The baselines were also implemented in PyTorch.

## 5.3 Evaluation Metrics

We use standard multiclass classification metrics: accuracy@K (ACC@K, K=1 and 5), macro precision (P), macro recall (R), and macro F1 score. ACC@K evaluates user-linking accuracy, while macro F1 provides a comprehensive evaluation across all classes, critical for imbalanced TUL datasets.

## 5.4 Overall Performance

We first present the results of our experiments on the Foursquare-NYC-Check-ins and Foursquare-TKY-Check-ins datasets in Tables 5 and 4. For the Foursquare-NYC-Check-ins dataset, GCN-TULHOR showed a noticeable yet

Table 4. Results on Foursqare-NYC-Check-ins. The highest performance is in bold, and the second highest is underlined. 'Improvement' shows the improvement of GCN-TULHOR over the strongest baseline.

| | **HO-Foursquare-NYC-Check-ins** | | | | | | | | | | | | | | |
| Model | $|\mathcal{U}| = 108$ | | | | | $|\mathcal{U}| = 209$ | | | | | $|\mathcal{U}| = 234$ | | | | |
| | Acc@1 | Acc@5 | P | R | F1 | Acc@1 | Acc@5 | P | R | F1 | Acc@1 | Acc@5 | P | R | F1 |
|---|---|---|---|---|---|---|---|---|---|---|---|---|---|---|---|
| DT | 0.884 | 0.892 | 0.878 | 0.867 | 0.868 | 0.785 | 0.788 | 0.753 | 0.728 | 0.730 | 0.778 | 0.782 | 0.722 | 0.712 | 0.705 |
| LDA | 0.822 | 0.851 | 0.920 | 0.810 | 0.868 | 0.746 | 0.781 | 0.791 | 0.687 | 0.713 | 0.696 | 0.752 | 0.724 | 0.615 | 0.650 |
| LINEAR-SVM | 0.873 | 0.929 | **0.966** | 0.878 | 0.909 | 0.776 | 0.839 | 0.785 | 0.702 | 0.727 | 0.731 | 0.798 | 0.724 | 0.628 | 0.657 |
| TULER | 0.870 | 0.929 | 0.869 | 0.851 | 0.852 | 0.776 | 0.853 | 0.749 | 0.722 | 0.718 | 0.768 | 0.844 | 0.733 | 0.707 | 0.703 |
| TULER-L | 0.903 | 0.942 | 0.904 | 0.890 | 0.890 | 0.847 | 0.898 | 0.828 | 0.803 | 0.807 | 0.845 | 0.889 | 0.821 | 0.806 | 0.803 |
| TULER-G | 0.909 | 0.949 | 0.914 | 0.897 | 0.898 | 0.854 | 0.892 | 0.835 | 0.811 | 0.812 | 0.846 | 0.891 | 0.821 | 0.805 | 0.803 |
| DeepTUL-LSTM | 0.823 | 0.896 | 0.715 | 0.703 | 0.709 | 0.716 | 0.832 | 0.554 | 0.559 | 0.556 | 0.712 | 0.830 | 0.569 | 0.557 | 0.563 |
| DeepTUL-GRU | 0.886 | 0.933 | 0.779 | 0.779 | 0.791 | 0.835 | 0.891 | 0.663 | 0.680 | 0.671 | 0.889 | **0.936** | 0.741 | 0.738 | 0.740 |
| DeepTul | 0.853 | 0.923 | 0.765 | 0.738 | 0.751 | 0.733 | 0.840 | 0.614 | 0.597 | 0.606 | 0.789 | 0.891 | 0.607 | 0.617 | 0.612 |
| TULHOR | <u>0.940</u> | <u>0.966</u> | 0.938 | <u>0.931</u> | <u>0.932</u> | <u>0.903</u> | <u>0.943</u> | <u>0.890</u> | <u>0.877</u> | <u>0.876</u> | <u>0.892</u> | 0.932 | <u>0.876</u> | <u>0.864</u> | <u>0.860</u> |
| GCN-TULHOR | **0.948** | **0.975** | 0.945 | **0.936** | **0.940** | **0.912** | **0.954** | **0.898** | **0.884** | **0.886** | **0.899** | **0.941** | **0.880** | **0.868** | **0.866** |
| Improvement | 0.85% | 0.93% | -2.17% | 0.54% | 0.86% | 0.99% | 1.16% | 0.89% | 0.80% | 1.00% | 0.78% | 0.53% | 0.46% | 0.46% | 0.69% |

Table 5. Results on Foursqare-TKY-Check-ins mobility dataset. The highest performance is indicated in bold and the second best performances has been underlined. 'Improvement' denotes the improvement of GCN-TULHOR model over the strongest baseline.

| | **HO-Foursquare-TKY-Check-ins** | | | | | | | | | | | | | | |
| Model | $|\mathcal{U}| = 108$ | | | | | $|\mathcal{U}| = 209$ | | | | | $|\mathcal{U}| = 451$ | | | | |
| | Acc@1 | Acc@5 | P | R | F1 | Acc@1 | Acc@5 | P | R | F1 | Acc@1 | Acc@5 | P | R | F1 |
|---|---|---|---|---|---|---|---|---|---|---|---|---|---|---|---|
| DT | 0.789 | 0.793 | 0.785 | 0.777 | 0.775 | 0.658 | 0.664 | 0.629 | 0.615 | 0.613 | 0.522 | 0.525 | 0.446 | 0.437 | 0.431 |
| LDA | 0.853 | 0.912 | 0.927 | 0.847 | 0.874 | 0.722 | 0.808 | 0.778 | 0.692 | 0.713 | 0.574 | 0.720 | 0.553 | 0.501 | 0.495 |
| LINEAR-SVM | 0.890 | 0.948 | 0.923 | 0.886 | 0.898 | 0.769 | 0.878 | 0.794 | 0.736 | 0.748 | 0.609 | 0.761 | 0.610 | 0.539 | 0.550 |
| TULER | 0.870 | 0.933 | 0.871 | 0.860 | 0.860 | 0.768 | 0.864 | 0.762 | 0.735 | 0.736 | 0.637 | 0.740 | 0.588 | 0.554 | 0.548 |
| TULER-L | 0.905 | 0.952 | 0.904 | 0.898 | 0.897 | 0.848 | 0.911 | 0.837 | 0.825 | 0.824 | 0.739 | 0.827 | 0.708 | 0.675 | 0.675 |
| TULER-G | 0.915 | 0.954 | 0.916 | 0.910 | 0.909 | 0.851 | 0.911 | 0.842 | 0.824 | 0.825 | 0.738 | 0.823 | 0.701 | 0.672 | 0.671 |
| DeepTUL-LSTM | 0.908 | 0.966 | 0.916 | 0.901 | 0.908 | 0.752 | 0.871 | 0.795 | 0.729 | 0.760 | 0.407 | 0.584 | 0.362 | 0.326 | 0.343 |
| DeepTUL-GRU | 0.933 | <u>0.975</u> | 0.932 | 0.928 | 0.930 | 0.869 | 0.937 | 0.872 | 0.856 | 0.864 | 0.742 | 0.821 | <u>0.715</u> | 0.689 | 0.695 |
| DeepTul | 0.922 | 0.966 | 0.927 | 0.913 | 0.920 | 0.773 | 0.904 | 0.820 | 0.747 | 0.782 | 0.660 | 0.790 | 0.631 | 0.587 | 0.608 |
| TULHOR | <u>0.939</u> | 0.973 | <u>0.937</u> | **0.934** | <u>0.933</u> | <u>0.893</u> | <u>0.953</u> | **0.883** | <u>0.877</u> | <u>0.875</u> | <u>0.801</u> | <u>0.888</u> | **0.783** | <u>0.755</u> | <u>0.752</u> |
| GCN-TULHOR | **0.945** | **0.976** | **0.938** | <u>0.932</u> | **0.935** | **0.894** | **0.956** | <u>0.882</u> | **0.878** | **0.876** | **0.802** | **0.889** | **0.783** | **0.756** | **0.753** |
| Improvement | 0.64% | 0.31% | 0.11% | -0.21% | 0.21% | 0.11% | 0.11% | 0.11% | 0.11% | 0.11% | 0.12% | 0.11% | 0.00% | 0.13% | 0.13% |

Table 6. Results on HO-Geolife, HO-Rome and HO-Porto mobility datasets. The highest performance is indicated in bold, and the second-best performance has been underlined. 'Improvement' denotes the improvement of GCN-TULHOR model over the strongest baseline.

| | **Impact of Models Across Continuous Datasets** | | | | | | | | | | | | | | |
| | **HO-Geolife** | | | | | **HO-Rome** | | | | | **HO-Porto** | | | | |
| Model | Acc@1 | Acc@5 | P | R | F1 | Acc@1 | Acc@5 | P | R | F1 | Acc@1 | Acc@5 | P | R | F1 |
|---|---|---|---|---|---|---|---|---|---|---|---|---|---|---|---|
| DT | 0.593 | 0.643 | 0.302 | 0.291 | 0.291 | 0.211 | 0.226 | 0.194 | 0.195 | 0.185 | 0.052 | <u>0.109</u> | 0.049 | 0.048 | 0.045 |
| LDA | 0.436 | 0.624 | 0.331 | 0.319 | 0.280 | 0.339 | 0.513 | 0.334 | 0.322 | 0.304 | 0.023 | 0.085 | 0.017 | 0.025 | 0.012 |
| LINEAR-SVM | 0.517 | 0.781 | 0.323 | 0.311 | 0.291 | 0.250 | 0.520 | 0.308 | 0.294 | 0.283 | 0.033 | 0.097 | 0.026 | 0.030 | 0.018 |
| RNN | 0.533 | 0.739 | 0.256 | 0.234 | 0.229 | 0.117 | 0.200 | 0.114 | 0.110 | 0.100 | 0.037 | 0.075 | 0.021 | 0.024 | 0.020 |
| LSTM | 0.594 | 0.812 | 0.306 | 0.299 | 0.287 | 0.124 | 0.237 | 0.117 | 0.120 | 0.112 | 0.039 | 0.084 | 0.024 | 0.028 | 0.023 |
| GRU | 0.599 | 0.823 | 0.315 | 0.297 | 0.294 | 0.111 | 0.204 | 0.126 | 0.109 | 0.103 | 0.041 | 0.089 | 0.027 | 0.031 | 0.025 |
| TULHOR | <u>0.630</u> | <u>0.870</u> | <u>0.420</u> | <u>0.430</u> | <u>0.430</u> | <u>0.400</u> | <u>0.550</u> | <u>0.370</u> | <u>0.380</u> | <u>0.360</u> | <u>0.090</u> | 0.074 | <u>0.070</u> | <u>0.090</u> | <u>0.070</u> |
| GCN-TULHOR | **0.670** | **0.870** | **0.440** | **0.480** | **0.470** | **0.420** | **0.620** | **0.390** | **0.400** | **0.370** | **0.100** | **0.160** | **0.080** | **0.100** | **0.080** |
| Improvement | 6.34% | 0.00% | 2.31% | 5.42% | 3.30% | 2.00% | 7.73% | 2.41% | 2.26% | 1.78% | 1.11% | 8.67% | 1.29% | 1.11% | 1.29% |

Table 7. Results on HO-Foursquare-TKY, HO-Foursquare-NYC, and HO-TDrive mobility datasets. The highest performance is indicated in bold, and the second-best performance has been underlined. 'Improvement' denotes the improvement of GCN-TULHOR model over the strongest baseline.

| | | | Impact of Models Across Continuous Datasets | | | | | | | | | | | |
| | **HO-TKY-CON** | | | | | **HO-NYC-CON** | | | | | **HO-TDrive** | | | | |
| Model | Acc@1 | Acc@5 | P | R | F1 | Acc@1 | Acc@5 | P | R | F1 | Acc@1 | Acc@5 | P | R | F1 |
|---|---|---|---|---|---|---|---|---|---|---|---|---|---|---|---|
| DT | 0.337 | 0.410 | 0.261 | 0.248 | 0.239 | 0.381 | 0.493 | 0.313 | 0.300 | 0.288 | 0.010 | 0.011 | 0.008 | 0.010 | 0.008 |
| LDA | 0.301 | 0.477 | 0.250 | 0.238 | 0.211 | 0.376 | 0.517 | 0.353 | 0.318 | 0.309 | 0.044 | 0.099 | 0.030 | 0.043 | 0.031 |
| LINEAR-SVM | 0.322 | 0.495 | **0.307** | 0.282 | 0.272 | 0.378 | 0.556 | 0.361 | 0.337 | **0.328** | 0.048 | 0.111 | 0.079 | 0.107 | 0.078 |
| RNN | 0.212 | 0.347 | 0.177 | 0.173 | 0.154 | 0.293 | 0.441 | 0.250 | 0.246 | 0.230 | 0.055 | 0.088 | 0.042 | 0.054 | 0.043 |
| LSTM | 0.267 | 0.415 | 0.221 | 0.219 | 0.200 | 0.326 | 0.481 | 0.280 | 0.278 | 0.257 | 0.052 | 0.094 | 0.040 | 0.051 | 0.040 |
| GRU | 0.270 | 0.420 | 0.222 | 0.221 | 0.201 | 0.334 | 0.487 | 0.272 | 0.281 | 0.257 | 0.057 | 0.088 | 0.045 | 0.053 | 0.043 |
| TULHOR | 0.342 | 0.510 | 0.284 | 0.274 | 0.262 | 0.397 | 0.567 | 0.350 | 0.335 | 0.312 | 0.090 | 0.074 | 0.070 | 0.090 | 0.070 |
| GCN-TULHOR | **0.358** | **0.536** | 0.298 | **0.288** | **0.274** | **0.417** | **0.595** | **0.368** | **0.352** | **0.328** | **0.100** | **0.160** | **0.080** | **0.100** | **0.080** |
| Improvement | 1.62% | 2.65% | 1.43% | 1.46% | 1.22% | 2.21% | 2.83% | 1.87% | 1.76% | 1.63% | 1.04% | 8.67% | 1.29% | 1.11% | 1.29% |

moderate improvement over TULHOR. For instance, in the $|\mathcal{U}| = 108$ user group, we observed an improvement in Acc@1 from 0.940 to 0.948, and Acc@5 from 0.966 to 0.975. Similarly, the F1 score improved from 0.932 to 0.940. For the Foursquare-TKY dataset, similar trends were observed. For example, in the $|\mathcal{U}| = 209$ user group, the Acc@1 increased from 0.893 to 0.894, while the F1 score improved slightly from 0.875 to 0.876.

These datasets, relying on sparse check-in events rather than continuous trajectories, face challenges like data sparsity and overfitting. While TULHOR has consistently achieved state-of-the-art performance, GCN-TULHOR shows meaningful and consistent improvements. These gains, observed across metrics like precision, recall, and F1, demonstrate GCN-TULHOR's ability to better capture spatial dependencies and generalize effectively without overfitting. By leveraging its graph convolutional architecture, GCN-TULHOR models richer relationships between user trajectories and POIs, surpassing TULHOR's capabilities and proving more robust in handling sparse, event-driven data.

## 5.5 Overall Performance on Continuous Data

Comparing the check-in and continuous data results shows a significant decline in performance. The decrease can be attributed to the inherent complexity of the problem. While continuous data reduces sparsity and increases the number of hexagon overlaps between trajectories, making the data richer, it also introduces additional challenges for learning. The larger dataset size, combined with the overlapping trajectories, makes it harder for the model to learn effective representations. This complexity arises from the higher dimensionality and the intricate nature of continuous movements, which require more sophisticated modeling to capture the true spatial relationships. This decrease in performance, however, better reflects the real-world challenges of modeling continuous trajectories, as they present a more accurate representation of user behavior compared to sparse check-in data.

Our results demonstrate that GCN-TULHOR consistently outperforms TULHOR and other baseline models across various continuous datasets, including HO-Geolife, HO-Rome, and HO-Porto, as shown in Table 6. For example, on the HO-Geolife dataset, GCN-TULHOR achieved an Acc@1 of 0.67, compared to 0.63 for TULHOR, and an F1 score of 0.47, compared to 0.43 for TULHOR. Similarly, on the HO-Rome dataset, GCN-TULHOR improved Acc@1 from 0.40 to 0.42 and F1 score from 0.36 to 0.37. These results highlight GCN-TULHOR's ability to effectively learn from raw continuous data, leveraging its graph convolutional architecture to capture nuanced spatial dependencies between trajectory points.

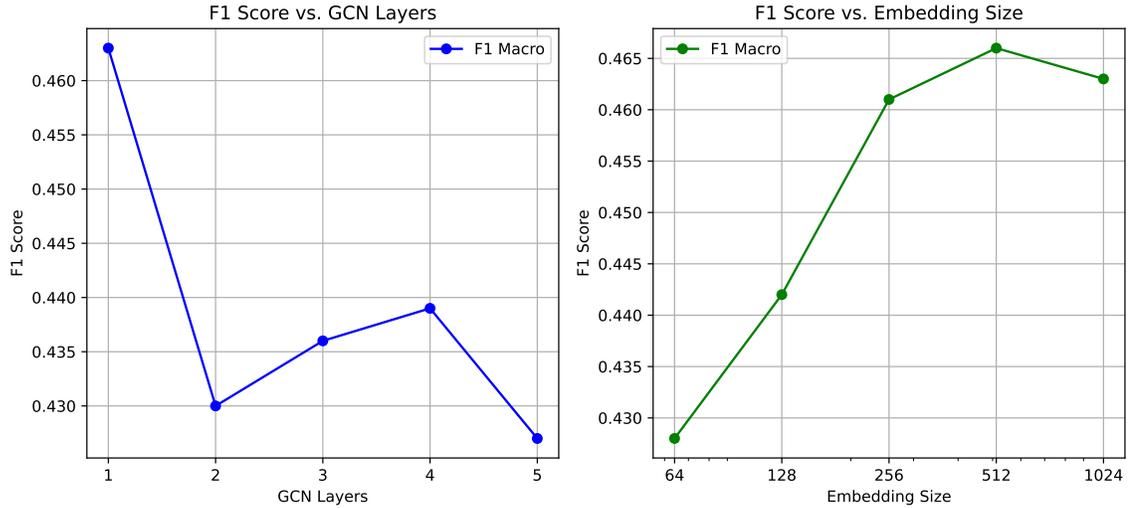

Fig. 8. Impact of varying the number of GCN layers and embedding dimensions on the Macro-F1 score of the GCN-TULHOR model using the HO-GᴇᴏLɪꜰᴇ dataset. This figure shows how performance initially peaks at a single GCN layer and then declines as more layers are added, and how the 512 embedding dimension performs the best.

The meaningful improvements achieved by GCN-TULHOR can be attributed to its inherent ability to process graph-structured data, which is particularly well-suited for continuous trajectories. It learns directly from the raw trajectory data, constructing a spatial graph to model interactions and dependencies between trajectory points. This allows GCN-TULHOR to generalize effectively across diverse datasets, avoiding overfitting while maintaining high performance. Especially when we rely on spatial data alone without using temporal or event-based features.

### 5.6 Sensitivity Study

**Impact of hyperparameters.** In this section, we analyze the impact of key hyperparameters unique to GCN-TULHOR, focusing on GCN layers and embedding size. Previous hyperparameter studies on TULHOR (e.g., embedding dimensions, hidden dimensions, attention heads, and layer counts) yielded consistent results when applied to GCN-TULHOR. Thus, we extend the study by examining GCN-specific parameters.

Performance evaluation, illustrated in Figure 8, reveals that a single GCN layer achieves the highest Macro-F1 score of 0.463. Adding more layers decreases performance, with five layers reducing the score to 0.427, likely due to over-smoothing, where node representations lose distinction—consistent with findings in graph neural network research. For embedding size, the model performs best at 512 dimensions, achieving a Macro-F1 score of 0.466. Larger embeddings, such as 1024 dimensions, result in a slight decline to 0.463, suggesting over-parameterization and diminishing returns due to increased complexity.

**Impact of Grid Cell Size on GCN-TULHOR Performance** We conducted an additional study to assess the impact of varying grid cell sizes on GCN-TULHOR's performance when processing continuous trajectory data. Similar to TULHOR, we tested resolutions HEX@k, where $k = \{6, 7, 8, 9, 10\}$. The smaller the $k$, the larger the cell size, resulting in fewer cells in the grid.

Table 8. Performance of GCN-TULHOR across different datasets and hexagon resolutions. Best values for each metric in each dataset are in bold.

| Dataset | HEX | Acc@1 | Acc@5 | P | R | F1 |
|---------|-----|-------|-------|---|---|-----|
| HO-Geolife | HEX6 | 0.477 | 0.744 | 0.236 | 0.306 | 0.242 |
| | HEX7 | 0.579 | 0.844 | 0.363 | 0.450 | 0.375 |
| | HEX8 | **0.668** | 0.874 | **0.431** | **0.470** | **0.440** |
| | HEX9 | 0.627 | **0.877** | 0.403 | 0.447 | 0.408 |
| | HEX10 | 0.549 | 0.804 | 0.322 | 0.334 | 0.318 |
| HO-TKY-CON | HEX6 | 0.149 | 0.306 | 0.117 | 0.112 | 0.093 |
| | HEX7 | 0.286 | 0.467 | 0.238 | 0.224 | 0.208 |
| | HEX8 | 0.358 | 0.536 | 0.298 | 0.288 | 0.274 |
| | HEX9 | **0.605** | **0.726** | **0.536** | **0.514** | **0.521** |
| | HEX10 | 0.493 | 0.619 | 0.424 | 0.409 | 0.408 |
| HO-NYC-CON | HEX6 | 0.168 | 0.345 | 0.134 | 0.137 | 0.115 |
| | HEX7 | 0.326 | 0.520 | 0.281 | 0.267 | 0.253 |
| | HEX8 | 0.417 | 0.595 | 0.368 | 0.352 | 0.328 |
| | HEX9 | **0.497** | **0.646** | **0.432** | **0.422** | **0.417** |
| | HEX10 | 0.465 | 0.568 | 0.407 | 0.391 | 0.388 |

The results of this experiment, summarized in Table 8, demonstrate that as the grid resolution increases (i.e., the cell size decreases), GCN-TULHOR's performance improves consistently across datasets. For *HO-NYC-CON*, GCN-TULHOR achieves its highest *F1 Macro* of 0.417 at HEX@9, outperforming HEX@7 and HEX@8 by 16% and 9%, respectively. A similar trend is observed in *HO-Geolife*, where the *F1 Macro* increases from 0.318 at HEX@10 to a peak of 0.440 at HEX@8. Interestingly, the accuracy and recall metrics also follow this pattern, underscoring the benefits of finer resolutions in capturing granular user movement patterns. However, excessively fine grids, such as HEX@10, introduce computational complexity and lead to a slight performance decline in datasets like *HO-TKY-CON*, where the *F1 Macro* decreases from 0.521 at HEX@9 to 0.407 at HEX@10. This suggests that HEX@9 offers a balance between granularity and computational feasibility for GCN-TULHOR.

The results highlight that finer grid resolutions allow GCN-TULHOR to better capture spatial dependencies in continuous data, enhancing its generalization across diverse datasets. Nevertheless, overly fine grids might negatively impact performance due to overfitting and computational overhead, emphasizing the importance of selecting an optimal resolution based on the dataset characteristics.

## 6 GENERALIZATION OF GCN SPATIAL EMBEDDINGS FOR TRAJECTORY MODELING

The integration of Graph Convolutional Networks (GCNs) into trajectory-user linking tasks demonstrates significant potential for learning spatial embeddings that capture nuanced relationships between trajectory points. These embeddings, constructed through hexagonal binning (HEX-level resolution), can be extracted from the GCN layer and utilized as additional features for models with lower capacity, such as LSTMs and RNNs. This approach enables these sequence-based models to leverage spatial insights derived from GCNs, improving their performance on tasks with high spatial complexity, and more importantly, improving generalization capabilities.

As shown in Table 9, we compare the performance of LSTM and GCN-LSTM, (LSTM with GCN embeddings) across various HEX resolutions on the HO-ROME dataset. The results illustrate that the inclusion of GCN-generated embeddings significantly enhances the generalization of sequential models. For instance, at HEX9, Acc@1 improves markedly from 0.1489 (LSTM) to 0.3896 (GCN-LSTM), while the F1 SCORE (MACRO) rises from 0.1431 to 0.3746. These improvements highlight the ability of GCN spatial embeddings to capture detailed spatial dependencies that sequential models struggle to learn independently, and to be effectively used by them.

Table 9. Performance of LSTM vs. GCN-LSTM on the ROME dataset at different HEX resolutions of HO-Rome. The highest performance is indicated in bold.

| HEX | LSTM | | | | | GCN-LSTM | | | | |
|---|---|---|---|---|---|---|---|---|---|---|
| | Acc@1 | Acc@5 | P | R | F1 | Acc@1 | Acc@5 | P | R | F1 |
| HEX6 | 0.1328 | **0.2962** | 0.1189 | 0.1251 | 0.1081 | 0.2503 | 0.4004 | 0.2153 | 0.2305 | 0.2182 |
| HEX7 | 0.1144 | 0.2648 | 0.1131 | 0.1089 | 0.1009 | 0.2978 | 0.4952 | 0.2801 | 0.2905 | 0.2756 |
| HEX8 | 0.1243 | 0.2371 | 0.1179 | 0.1206 | 0.1126 | 0.3431 | 0.5299 | 0.3157 | 0.3300 | 0.3188 |
| HEX9 | **0.1489** | 0.2295 | **0.1679** | **0.1441** | **0.1431** | **0.3896** | **0.5557** | **0.3727** | **0.3869** | **0.3746** |
| HEX10 | 0.1328 | 0.2018 | 0.1196 | 0.1249 | 0.1050 | 0.3687 | 0.5158 | 0.3502 | 0.3606 | 0.3488 |

## 7 RELATED WORK

The TUL task, associating anonymized trajectories with their generating users, has gained considerable interest due to its relevance to personalized recommendations, privacy preservation, urban planning, and secure location-based services. This section reviews the literature across several interconnected fields: foundational trajectory similarity measures, trajectory representation learning, deep learning advancements, graph-based mobility modeling, spatial tessellation, and recent methodological developments.

### 7.1 Foundational Trajectory Similarity Measures

Early trajectory analysis focused on defining robust similarity measures for trajectory comparison, foundational for retrieval, clustering, and classification. Dynamic Time Warping (DTW) [3] was initially proposed for time series alignment, later adapted to trajectory comparison due to its robustness against variations in speed and sampling rates. Similarly, Longest Common Subsequence (LCSS) [22] identifies shared subsequences within trajectories, and Edit Distance on Real Sequences (EDR) [6] calculates minimal edit operations to transform one trajectory into another, demonstrating resilience to spatial noise. Fréchet distance and its discrete counterpart [7] provide geometric trajectory similarity measures. Despite their utility, these methods are computationally intensive, spurring research towards more scalable representation techniques [32].

### 7.2 Trajectory Representation Learning (TRL)

TRL emerged to address computational inefficiencies, embedding trajectories into low-dimensional spaces facilitating efficient similarity comparisons and downstream analyses [26–28]. This transition toward learned representations significantly impacted subsequent developments in deep learning for TUL.

### 7.3 Deep Learning and Probabilistic Approaches

Initial applications of probabilistic models, notably Markov Chains (MCs), laid the groundwork for trajectory-user linking and prediction [2, 17]. However, MCs' assumption of immediate-state dependency limits modeling of complex mobility patterns. Deep learning's advent, particularly Recurrent Neural Networks (RNNs), Long Short-Term Memory (LSTM), and Gated Recurrent Units (GRUs), addressed these limitations by effectively modeling sequential data [9]. Gao et al.'s TULER demonstrated RNN effectiveness in embedding trajectory sequences [9].

### 7.4 Advanced Deep Learning Techniques

Variational Autoencoders (VAEs) improved sparse data handling by leveraging unlabeled trajectories to learn robust generative representations [36]. Attention mechanisms, derived from natural language processing, further enhanced modeling capabilities, allowing effective capture of salient trajectory points and long-range dependencies [14, 18, 20]. DeepTUL employed recurrent networks with attention to address sparsity and higher-order mobility patterns [14].

### 7.5 Graph Neural Networks (GNNs) and Mobility Modeling

Recognizing the limitations of purely sequential models in spatial modeling, recent approaches have utilized Graph Neural Networks (GNNs). Zhou et al. proposed GNNTUL, exploiting spatial graphs to effectively represent mobility patterns [37]. Sun et al.'s AttnTUL combined GNNs with hierarchical attention to encode both global mobility graphs and local sequences [19]. Chang et al.'s HGTUL employed hypergraph neural networks for modeling higher-order co-occurrences within trajectories [5], enhancing the expressiveness of spatial relationships.

### 7.6 Spatial Tessellation and Higher-Order Encoding

Spatial abstraction techniques, notably hexagonal tessellation as demonstrated by Point2Hex [8], effectively address trajectory sparsity by mapping raw coordinates into structured grid cells. TULHOR utilized higher-order spatial abstractions, abstracting mobility into sequences of hexagonal cells to reduce sparsity and capture underlying mobility semantics [1]. This approach has shown notable effectiveness in enhancing trajectory representation for sparse check-in data.

### 7.7 Sparse vs. Continuous Trajectory Data

Research often distinguishes between sparse check-in data, typically event-based, and continuous GPS trajectory data, captured at regular intervals. Sparse data methods focus heavily on discrete events and POI semantics [2, 24]. In contrast, continuous trajectory methods must handle detailed spatiotemporal dynamics and route nuances [29, 34]. Models effective in sparse scenarios often generalize poorly to continuous data due to fundamental differences in data generation and mobility representation.

### 7.8 Recent Trends and Large Models

Recent methodological advancements include semi-supervised learning (SSL) frameworks, such as TULVAE [36], and large-scale language models (LLMs) adapted for mobility tasks (e.g., Mobility-LLM [10]). SSL leverages unlabeled data to improve modeling robustness, while LLMs offer sophisticated semantic understanding capabilities that are being increasingly applied to trajectory prediction and user linking tasks.

Our work, GCN-TULHOR, builds directly upon TULHOR's foundation, integrating Graph Convolutional Networks (GCNs) to explicitly model complex spatial dependencies and graph-structured movement patterns. Unlike prior models that primarily operate on linear trajectory sequences or discrete events, GCN-TULHOR introduces a unified higher-order mobility flow representation that accommodates both sparse and continuous trajectories. By discretizing trajectories into hexagonal grid cells, our approach harmonizes spatial abstraction with explicit spatial graph modeling, enhancing representational power and addressing key limitations identified in current methods [1, 8].

In conclusion, by combining advanced deep learning, higher-order spatial abstractions, and graph-based spatial modeling, GCN-TULHOR offers a robust, scalable, and generalized framework for trajectory-user linking, demonstrating superior performance across diverse mobility scenarios.

## 8 CONCLUSIONS

In this paper, we introduced GCN-TULHOR, an advanced TUL model that enhances the existing TULHOR framework by integrating GCNs. Our approach effectively leverages higher-order mobility flow representations and explicit graph-based spatial modeling to robustly link trajectories to their respective users, outperforming existing baselines across diverse datasets. By learning spatial embeddings directly from trajectory data, without relying on auxiliary side information such as timestamps or POIs, GCN-TULHOR significantly improves its applicability, particularly in privacy-sensitive and data-scarce scenarios.

The comprehensive evaluation across six diverse datasets—including sparse check-in datasets like Foursquare-NYC-Check-ins and Foursquare-TKY-Check-ins, and continuous trajectory datasets such as HO-T-Drive, HO-Porto, HO-Rome, and HO-Geolife—confirms the robustness of our approach. GCN-TULHOR consistently achieved state-of-the-art performance, demonstrating marked improvements in key metrics such as accuracy, precision, recall, and F1-score. The sensitivity studies highlighted that the model's optimal configuration typically involves a single GCN layer and an embedding dimension of 512, balancing model complexity and expressive power effectively.

Furthermore, the unified higher-order mobility flow representation via hexagonal tessellation significantly reduced data sparsity and enhanced the semantic understanding of user mobility patterns. The learned GCN spatial embeddings proved valuable not only for improving trajectory-user linking but also for augmenting the capabilities of other sequence-based models, indicating strong transferability of spatial knowledge.

**Limitations**

Although GCN-TULHOR generally outperforms other methods, it can sometimes achieve only marginal improvements, performing on par with baseline methods in certain scenarios. A primary limitation is the model's increased computational complexity, which leads to longer training and inference times. This issue is especially pronounced with very large datasets and extremely fine-grained spatial tessellations, limiting the model's scalability.

To address these limitations, future work could explore advanced GNN architectures such as Graph Attention Networks (GAT) [21] and Graph Isomorphism Networks (GIN) [23] to further enhance the model's spatial representations. Integrating temporal components into the framework (for example, through spatiotemporal GNNs or temporal attention mechanisms [31]) is another promising direction to capture time-dependent mobility patterns. Finally, future efforts will focus on optimizing the model's computational efficiency, potentially through textual or spatial encoders and indexing strategies, to ensure that GCN-TULHOR can scale effectively while maintaining its TUL capabilities.

## DATA AND CODE AVAILABILITY

To foster transparency, reproducibility, and further research, we have made the complete source code for our experiments publicly accessible at: https://github.com/pranavgupta0001/GCN-TULHOR[2]. All datasets used for training

---

and evaluation are entirely open-source and have been appropriately cited within the manuscript. No proprietary or restricted-access data has been used.

## ETHICS STATEMENT

Our research utilizes real-world trajectory datasets, which may raise concerns about individual privacy and the potential for re-identification. Therefore, to ensure the protection of privacy and uphold ethical considerations, all datasets used in our evaluation have been anonymized, meaning that personally identifiable information such as names, addresses, and other explicit identifiers have been removed by the original data providers before being made publicly available. These datasets are derived from publicly available sources, which have been curated to ensure free use for research purposes, as outlined in their terms of use. We have strictly adhered to all terms and conditions associated with the use of these datasets, and proper attribution has been given to the original data providers through citations within this paper. Our analysis does not involve any sensitive or private information beyond the spatiotemporal data already anonymized by the original providers, and our study primarily focuses on analyzing aggregate mobility patterns without attempting to identify or track specific individuals. The main goal of our work is to improve algorithmic approaches to trajectory-user linking rather than to compromise the privacy of individuals. The authors declare no competing interests.

## ACKNOWLEDGEMENTS

K.T. and P.G. designed the project and executed all experimental procedures. The manuscript was collaboratively authored by P.G., K.T., and M.P., with M.P. providing comprehensive supervision throughout the project.